\documentclass[a4paper, conference]{IEEEtran}

\usepackage{graphicx, acronym, ifthen, substr}

\newcommand{\tfigure}[9]
	{
	\IfSubStringInString{!}{#7}{\begin{figure}[#7]}{\IfSubStringInString{*}{#7}{\begin{figure*}[!t]}{\begin{figure}[!t]}}
	\IfSubStringInString{mm}{#8}{\vspace{#8}}{}

	\centering
	\IfSubStringInString{pdf}{#3}
		{\includegraphics[#1]{images/#2}}
		{\includegraphics[#1]{images/#2-crop.pdf}}

	\IfSubStringInString{m}{#6}{\vspace{#6}}{}
	\caption{\label{#2}#4: #5}
	\IfSubStringInString{mm}{#9}{\vspace{#9}}{}
	\IfSubStringInString{*}{#7}{\end{figure*}}{\end{figure}}
	}

\newcommand{\execute}[1]{\immediate\write18{#1}}

\begin{document}

\title{Ecosystem-Oriented\\Distributed Evolutionary Computing}

\author{\IEEEauthorblockN{Gerard Briscoe}
\IEEEauthorblockA{Systems Research Group\\
Computer Laboratory\\
University of Cambridge\\
Email: gerard.briscoe@cl.cam.ac.uk}
\and
\IEEEauthorblockN{Philippe De Wilde}
\IEEEauthorblockA{Intelligent Systems Lab\\
School of Mathematical and Computer Sciences\\
Heriot Watt University\\
Email: p.de\_wilde@hw.ac.uk}}

\maketitle

\begin{abstract}
We create a novel optimisation technique inspired by natural ecosystems, where the optimisation works at two levels: a first optimisation, migration of genes which are distributed in a peer-to-peer network, operating continuously in time; this process feeds a second optimisation based on evolutionary computing that operates locally on single peers and is aimed at finding solutions to satisfy locally relevant constraints. We consider from the domain of computer science distributed evolutionary computing, with the relevant theory from the domain of theoretical biology, including the fields of evolutionary and ecological theory, the topological structure of ecosystems, and evolutionary processes within distributed environments. We then define ecosystem-oriented distributed evolutionary computing, imbibed with the properties of self-organisation, scalability and sustainability from natural ecosystems, including a novel form of \acl{DEC}. Finally, we conclude with a discussion of the apparent compromises resulting from the hybrid model created, such as the network topology.

\end{abstract}

\acrodef{PCG}{Projected Conjugate Gradient} 
\acrodef{QP}{quadratic programming}
\acrodef{RBF}{Radial-Basis Function}
\acrodef{ABM}{Agent-Based Modelling}
\acrodef{AI}{Artificial Intelligence}
\acrodef{DAI}{Distributed Artificial Intelligence}
\acrodef{API}{Application Programming Interface}
\acrodef{ARF}{p14ARF human tumor-suppressor gene}
\acrodef{B2B}{business-to-business}
\acrodef{BDP}{Biological Design Pattern}
\acrodef{BGS}{Best Guess Solution}
\acrodef{BIC}{Biologically-Inspired Computing}
\acrodef{BML}{Business Modelling Language}
\acrodef{BPEL}{Business Process Execution Language}
\acrodef{BPMN}{Business Process Modelling Notation}
\acrodef{CAS}{Complex Adaptive Systems}
\acrodef{COBOL}{COmmon Business-Oriented Language}
\acrodef{DBE}{Digital Business Ecosystem}
\acrodef{DE}{Digital Ecosystem}
\acrodef{DEC}{distributed evolutionary computing}
\acrodef{DGA}{Distributed genetic algorithms}
\acrodef{DIS}{Distributed Intelligence System}
\acrodef{DNA}{Deoxyribose Nucleic Acid}
\acrodef{DOP}{DBE Open Protocol}
\acrodef{DSS}{Distributed Storage System}
\acrodef{EAP}{Evolving Agent Population}
\acrodef{ebXML}{e-business eXtensible Markup Language}
\acrodef{EC}{Evolutionary Computing}
\acrodef{ECJ}{Evolutionary Computing in Java}
\acrodef{EE}{Evolutionary Environment}
\acrodef{EFL}{Evolutionary Framework for Language}
\acrodef{FLE}{Framework for Language Ecosystems}
\acrodef{EOA}{Ecosystem-Oriented Architecture}
\acrodef{ESS}{evolutionary stable strategy}
\acrodef{EvE}{Evolutionary Environment}
\acrodef{ExE}{Execution Environment}
\acrodef{FCB}{Framework for Computational Biomimicry}
\acrodef{FFF}{Fitness Function Framework}
\acrodef{FL}{Fitness Landscape}
\acrodef{HWU}{Heriot-Watt University}
\acrodef{ICL}{Imperial College London}
\acrodef{ICT}{Information and Communications Technology}
\acrodef{INTEL}{Intel Ireland}
\acrodef{IPA}{International Phonetic Alphabet}
\acrodef{ISUFI}{Istituto Superiore Universitario di Formazione Interdisciplinare}
\acrodef{JDJ}{Java Developer's Journal}
\acrodef{KC}{Kolmogorov-Chaitin}
\acrodef{LAN}{local area network}
\acrodef{LSE}{London School of Economics and Political Science}
\acrodef{MAS}{Multi-Agent System}
\acrodef{MDL}{Minimum Description Length}
\acrodef{MDM2}{murine double minute 2}
\acrodef{MFT}{Mean Field Theory}
\acrodef{MoAS}{Mobile Agent System}
\acrodef{MOF}{Meta Object Facility}
\acrodef{MUH}{migration and usage history}
\acrodef{NIC}{Nature Inspired Computing}
\acrodef{NN}{Neural Network}
\acrodef{NoE}{Network of Excellence}
\acrodef{OMG}{Open Mac Grid}
\acrodef{OPAALS}{Open Philosophies for Associative Autopoietic Digital Ecosystems}
\acrodef{P2P}{peer-to-peer}
\acrodef{P53}{protein 53}
\acrodef{PDA}{Personal Digital Assistant}
\acrodef{QoS}{quality of service}
\acrodef{REST}{REpresentational State Transfer}
\acrodef{RNA}{Deoxyribose Nucleic Acid}
\acrodef{SAE}{Software Agent Ecosystem}
\acrodef{SBML}{Systems Biology Modelling Language}
\acrodef{SBVR}{Semantics of Business Vocabulary and Business Rules}
\acrodef{SDL}{Service Description Language}
\acrodef{SF}{Service Factory}
\acrodef{SIM}{Social Interaction Mechanism}
\acrodef{SM}{Service Manifest}
\acrodef{SME}{Small and Medium sized Enterprise}
\acrodef{SML}{Service Modelling Language}
\acrodef{SMO}{Sequential Minimal Optimisation}
\acrodef{SOA}{Service-Oriented Architecture}
\acrodef{SOAP}{Simple Object Access Protocol}
\acrodef{SOC}{Self-Organised Criticality}
\acrodef{SOLUTA}{SOLUTA.NET}
\acrodef{SOM}{Self-Organising Map}
\acrodef{SSL}{Semantic Service Language}
\acrodef{STU}{Salzburg Technical University}
\acrodef{SUN}{Sun Microsystems}
\acrodef{SVM}{Support Vector Machine}
\acrodef{TM}{Turing Machine}
\acrodef{UBHAM}{University of Birmingham}
\acrodef{UDDI}{Universal Description Discovery and Integration}
\acrodef{UML}{Unified Modelling Language}
\acrodef{URI}{Uniform Resource Identifier}
\acrodef{UTM}{Universal Turing Machine}
\acrodef{VLP}{variable length population}
\acrodef{VLS}{variable length sequences}
\acrodef{vls}{variable length sequence}
\acrodef{WP}{Work-Package}
\acrodef{WSDL}{Web Services Definition Language}
\acrodef{XMI}{XML Metadata Interchange}
\acrodef{XML}{eXtensible Markup Language}
\acrodef{MD5}{Message-Digest algorithm 5}
\acrodef{GA}{genetic algorithm}
\acrodef{GP}{genetic programming}
\acrodef{MASON}{Multi-Agent Simulator Of Neighbourhoods}
\acrodef{Repast}{Recursive Porous Agent Simulation Toolkit}
\acrodef{JCLEC}{Java Computing Library for Evolutionary Computing}
\acrodef{OWL-S}{Web Ontology Language - Service}
\acrodef{EGT}{Evolutionary Game Theory}
\acrodef{RBF}{Radial Basis Functions}
\acrodef{SWS}{Semantic Web Services}
\acrodef{HDD}{Hard Disk Drive}
\acrodef{SSD}{Solid-State Drive}

\execute{echo > captions.tex}

\section{Introduction}

In \acl{BIC}, one of the primary sources of inspiration from nature has been evolution \cite{ecpaper}. Evolution has been clearly identified as the source of many diverse and creative solutions to problems in nature \cite{ec15, ec16}. However, it can also be useful as a problem-solving tool in artificial systems. Computer scientists and other theoreticians realised that the selection and mutation mechanisms that appear so effective in biological evolution could be abstracted to be implemented in a computational algorithm \cite{ecpaper}. Evolutionary computing is now recognised as a sub-field of artificial intelligence (more particularly computational intelligence) that involves combinatorial optimisation problems \cite{ec17}.

The use of \emph{evolution} for algorithm generation may be well understood within computer science under the auspices of \emph{evolutionary computing} \cite{eiben2003iec}, but \emph{ecology} is not. Ecosystems are considered to be robust, scalable architectures that can automatically solve complex, dynamic problems, possessing several properties that may be useful in automated systems. These properties include self-organisation, self-management, scalability, the ability to provide complex solutions, and the automated composition of these complex solutions \cite{Levin}. So, potential exists to create an ecosystem-oriented form of evolutionary computing. We propose that an ecosystem inspired approach, would be more effective at greater scales than traditionally inspired approaches, because it would be built upon the scalable and self-organising properties of ecosystems \cite{Levin}.
 
In this paper, we give a brief introduction to the theory of distributed evolutionary computing (section \ref{dec}). We then introduce ecosystems, including relevant issues around networks and spacial dynamics (section \ref{theory}). We then describe what is required for ecosystem-oriented distributed evolution (section \ref{vision}), before illustrating our model of ecosystem-oriented distributed evolutionary computing (section \ref{compModel}). We then consider its emergent topology (section \ref{topology}), before then discussing the apparent compromises resulting from the hybrid model created (section \ref{discussion}), and stating our conclusions (section \ref{conclusion}).

\section{Distributed Evolutionary Computing}
\label{dec}

The fact that evolutionary computing manipulates a population of independent solutions actually makes it well suited for parallel and distributed computation architectures \cite{cantupaz1998spg}, i.e. evolution is well suited to the ecological spaces in which it occurs. The motivation for using parallel or distributed evolutionary algorithms is twofold: first, improving the speed of evolutionary processes by conducting concurrent evaluations of individuals in a population; second, improving the problem-solving process by overcoming difficulties that face traditional evolutionary algorithms, such as maintaining diversity to avoid premature convergence \cite{stender1993pga}. There are several variants of distributed evolutionary computing, with there being two main forms \cite{cantupaz1998spg,stender1993pga}:

Fine-grained \emph{diffusion} models \cite{stender1993pga} assign one individual per processor. A local neighbourhood topology is assumed, and individuals are allowed to mate only within their neighbourhood, called a \emph{deme}\footnote{In biology a deme is a term for a local population of organisms of one species that actively interbreed with one another and share a distinct gene-pool \cite{devisser2007ese}.}. The demes overlap by an amount that depends on their shape and size, and in this way create an implicit migration mechanism. Each processor runs an identical evolutionary algorithm which selects parents from the local neighbourhood, produces an offspring, and decides whether to replace the current individual with an offspring. However, even with the advent of multi-processor computers, multi-core processors, and large scale distributed computing (cloud computing) which provide the ability to execute multiple threads simultaneously, this approach would still prove impractical in supporting the number of populations necessary to create an ecosystem-orientated form of distributed evolutionary computing.

In the coarse-grained \emph{island} models \cite{lin1994cgp,cantupaz1998spg}, evolution occurs in multiple parallel sub-populations (islands), each running a local evolutionary algorithm, evolving independently with occasional \emph{migrations} of highly fit individuals among sub-populations. An example island-model \cite{lin1994cgp,cantupaz1998spg} is visualised in Figure \ref{islandModel}, in which there are different probabilities of going from island 1 to island 2, as there is of going from island 2 to island 1. This allows maximum flexibility for the migration process, and mirrors the naturally inspired quality that although two populations have the same physical separation, it may be easier to migrate in one direction than the other, i.e. fish migration is easier downstream than upstream. The migration of the \emph{island} models is like the notion of migration in nature, being similar to the metapopulation models of theoretical ecology \cite{levins1969sda}. However, all the \emph{islands} in this approach work on exactly the same problem, which makes it less analogous to ecosystems in which different locations can be environmentally different \cite{begon96}. We will take advantage of this property later when defining ecosystem-orientated distributed evolutionary computing.

\tfigure{width=0.7\columnwidth}{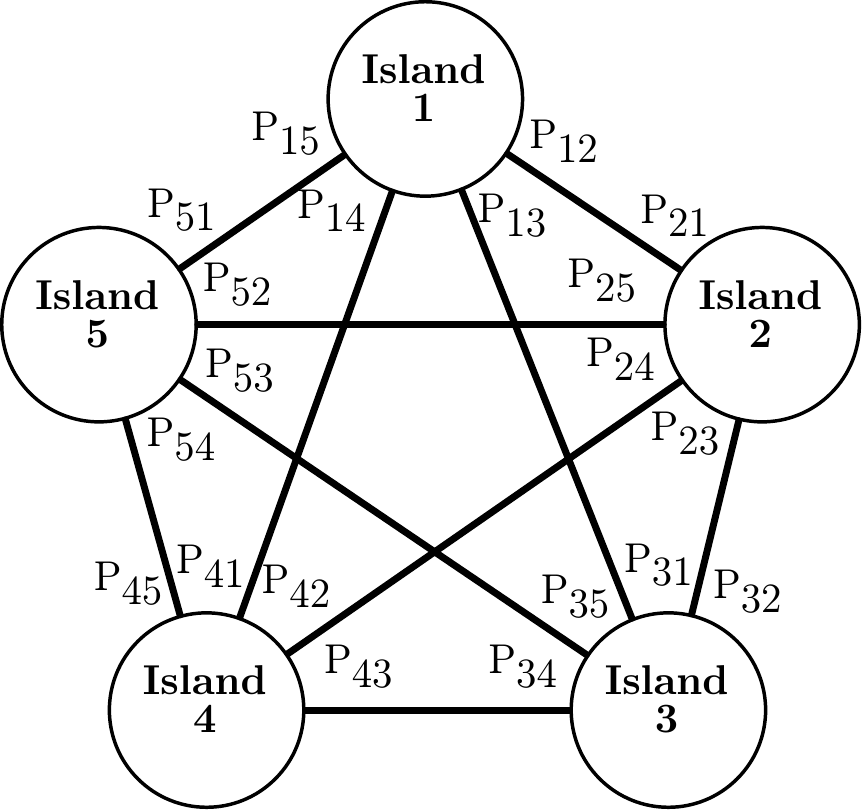}{graffle}{Island-Model of Distributed Evolutionary Computing}{There are different probabilities of going from island 1 to island 2, as there is of going from island 2 to island 1.}{-2mm}{}{}{}

\subsection{Other Work}

While there have been past attempts to have ecologically-inspired forms of evolutionary computing, for example \cite{yuchi2008,hoile2002core}, they have all lacked actual ecological dynamics. In the case of \cite{yuchi2008} they attempt to include a feature of population dynamics, which they incorrectly refer to as ecology-inspired. While we have previously considered \cite{epi,acmMedes,thesis,dbebkpub,de07oz,bionetics} the possibility of ecosystem-oriented \ac{DEC}, but only in very specific applications, so now we can consider it abstractly from which it can wide applicability.

\section{Natural Ecosystems}
\label{theory}

\tfigure{width=0.9\columnwidth}{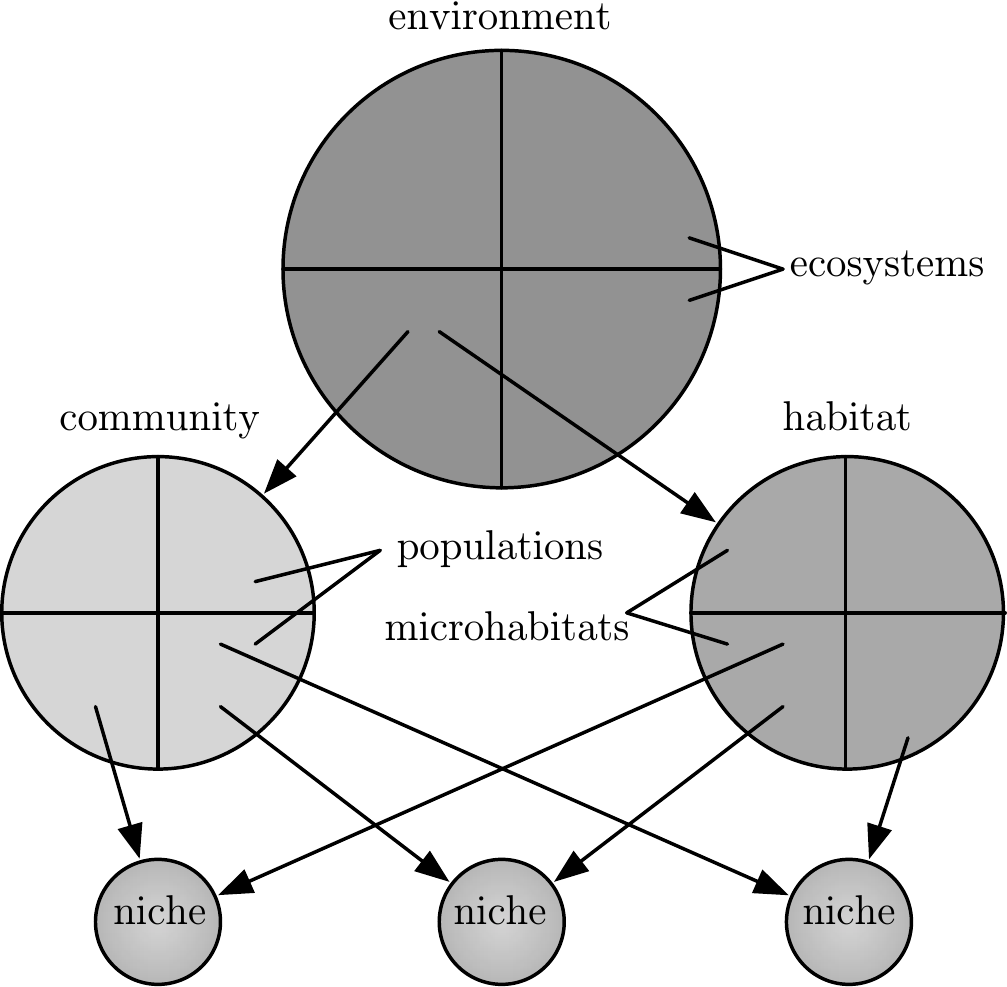}{graffle}{Ecosystem Structure}{(redrawn from \cite{longman}): A stable, self-perpetuating system made up of one or more communities of organisms, consisting of species in their habitats, with their populations existing in their respective micro-habitats \cite{begon96}.}{-2mm}{!b}{}{}

An ecosystem is a natural unit made up of living (biotic) and non-living (abiotic) components, from whose interactions emerge a dynamically stable, self-perpetuating system. It is made up of one or more communities of organisms, consisting of species in their habitats, with their populations existing in their respective micro-habitats \cite{begon96}. A community is a naturally occurring group of populations from different species that live together, and interact as a self-contained unit in the same habitat. A habitat is a distinct part of the environment \cite{begon96}, for example, a stream. Individual organisms migrate through the ecosystem into different habitats competing with other organisms for limited resources, with a population being the aggregate number of the individuals, of a particular species, inhabiting a specific habitat or micro-habitat \cite{begon96}. A micro-habitat is a subdivision of a habitat that possesses its own unique properties, such as a micro-climate \cite{lawrence1989hsd}. Evolution occurs to all living components of an ecosystem, with the evolutionary pressures varying from one population to the next depending on the environment that is the population's habitat. A population, in its micro-habitat, comes to occupy a niche, which is the functional relationship of a population to the environment that it occupies. A niche results in the highly specialised adaptation of a population to its micro-habitat \cite{lawrence1989hsd}.

\subsection{Networks and Spatial Dynamics}

A key factor in the maintenance of diversity in ecosystems is spatial interactions, and several modelling systems have been used to represent these spatial interactions, including metapopulations\footnote{A metapopulation is a collection of relatively isolated, spatially distributed, local populations bound together by occasional dispersal between populations \cite{levins1969sda,hanski1999me,hanski2003mtf}.}, diffusion models, cellular automata and agent-based models (termed individual-based models in ecology) \cite{Greenetal2006}. The broad predictions of these diverse models are in good agreement. At local scales, spatial interactions favour relatively abundant species disproportionately. However, at a wider scale, this effect can preserve diversity, because different species will be locally abundant in different places. The result is that even in homogeneous environments, population distributions tend to form discrete, long-lasting patches that can resist an invasion by superior competitors \cite{Greenetal2006}. Population distributions can also be influenced by environmental variations such as barriers, gradients, and patches. The possible behaviour of spatially distributed ecosystems is so diverse that scenario-specific modelling is necessary to understand any real system \cite{suzie}. Nonetheless, certain robust patterns are observed. These include the relative abundance of species, which consistently follows a roughly log-normal relationship \cite{Bell}, and the relationship between geographic area and the number of species present, which follows a power law \cite{sizling2004pls}. The reasons for these patterns are disputed, because they can be generated by both spatial extensions of simple Lotka-Volterra competition models \cite{Hubbell}, and more complex ecosystem models \cite{Sole}. 

In a digital context, we can consider spatial interactions as those that can arise within distributed systems, a set of interconnected locations, with entities that can migrate between these connected locations. In such systems the spatial dynamics are relatively simple compared with those seen in ecosystems, which incorporate barriers, gradients, and patchy environments at multiple scales in continuous space \cite{begon96}. Nevertheless, depending on how the connections between locations are organised such synthetic systems might have dynamics that closely parallel to spatially explicit models, diffusion models, or metapopulations \cite{suzie}.

From an information theory perspective, this change in landscape connectivity can mediate global and local search strategies \cite{Greenetal2000}. In a well-connected landscape, selection favours the globally superior, and pursuit of different evolutionary paths is discouraged, potentially leading to premature convergence. When the landscape is fragmented, populations may diverge, solving the same problems in different ways. It has been suggested that the evolution of complexity in nature involves repeated landscape phase changes, allowing selection to alternate between local and global search \cite{Greenetalinpress}. So, we can potentially take advantage of this by using a diverse heterogeneous distributed landscape (topology), i.e. a distributed environment, for our ecosystem-oriented distributed evolutionary computing.

\tfigure{width=0.9\textwidth}{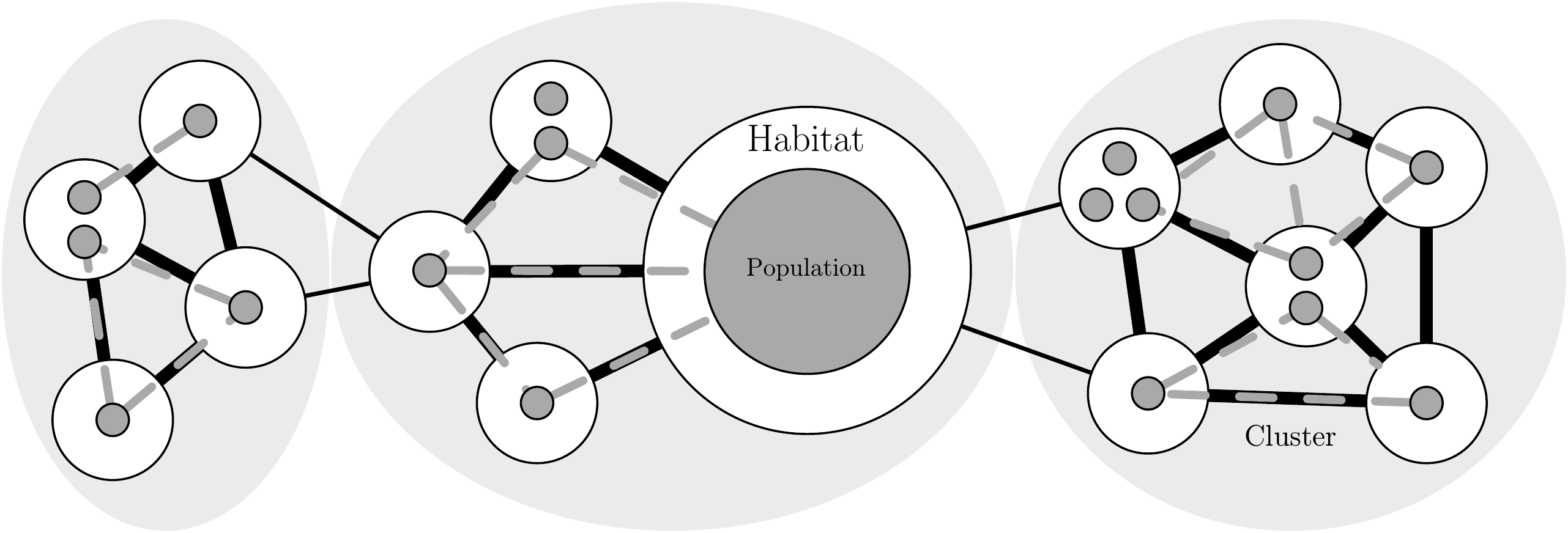}{graffle}{Distributed Evolution in Ecosystem-Oriented Distributed Evolutionary Computing}{Different requests are evaluated on separate \emph{islands} (populations), with their evolution accelerated by the sharing of solutions between the evolving populations (islands), because they are working to solve similar requests (problems). The yellow lines connecting the evolving populations indicate similarity in the requests being managed.}{-2mm}{*}{}{-3mm}

\section{Ecosystem-Oriented Distributed Evolution}
\label{vision}
Our ecosystem-oriented form of distributed evolutionary computing is uniquely defined by the creation of multiple evolving populations in response to \emph{similar} requests, whereas in the island-models of \acl{DEC} there are multiple evolving populations in response to only one request \cite{lin1994cgp}. So, different requests are evaluated on separate \emph{islands} (populations), with their evolution accelerated by the sharing of solutions between the evolving populations (islands), because they are working to solve similar requests (problems). Furthermore, the islands (habitats) will be able to support multiple populations, like a natural ecosystem. This is shown in Figure \ref{similar}, where the dashed yellow lines connecting the evolving populations indicate similarity in the requests being managed.

Our distributed evolutionary computing architecture favours the use of Pareto-sets for fitness determination, because Pareto optimisation for multi-objective problems is usually most effective with spatial distribution of the populations, as partial solutions (solutions to different niches) evolve in different parts of a \emph{distributed population} \cite{detoro2002ppg} (i.e. different populations in different habitats). By contrast, in a single population, individuals are always interacting with each other, via crossover, which does not allow for this type of specialisation \cite{back1996eat}.

It should be noted that our approach requires a sufficiently large user base, so that there can be communities within the user base, and therefore allow for similarity in the user requests. Assuming a user base of hundreds of users, then there would be hundreds of habitats, in which there will be potentially three or more times the number of populations at any one time. Then there will be thousands of genes (atomic algorithms) available to meet the requests for algorithms from the users. In such a scenario, there would be a sufficient number of users to find similarity within their requests, and therefore apply our novel form of \acl{DEC}.

\section{Computational Model}
\label{compModel}

\subsection{Gene}

The gene, $g$, represents the base unit of the evolutionary computing processes, in the same way that genes are the functional unit in evolution \cite{lawrence1989hsd}, and so are atomic (irreducible) algorithms. Collections of genes are functionally analogous to the organisms of ecosystems, including their behaviour of migration \cite{begon96}. While \emph{evolutionary computing} \cite{eiben2003iec} will be used for \emph{combinatorial optimisation} \cite{papadimitriou1998coa} to evolve optimal set of genes, $\{g_{i}, ..., g_{j}\}$, in response to a user $u$ request $r$ for an algorithm. So, the genes $g$ will migrate through the habitat network ($H=\{h_{1}, ..., h_{i}\}$) adapting to find \emph{niches} where they are useful in fulfilling other user requests for algorithms $r$. Therefore, interacting and evolving, adapting over time $t$ to the environment, and thereby serving the ever-changing requirements imposed by a user base $U=\{u_{1}, ..., u_{i}\}$.

The genes $g$ migrate through the interconnected habitats $h$ combining with one another in populations $P$ to meet user requests for algorithms $r$. The migration path from the current habitat $h$ is dependent on the connections $c$ between the habitats $h$, the \emph{migration probabilities} $c=\{p(h_{i}), ..., p(h_{j})\}$. The migration of a gene $g$ is initially triggered by deployment to its user's habitat $h$, for distribution to other users who will potentially make use of the algorithm that the gene $g$ represents. So, upon deployment a gene $g$ is copied to the gene-pool\footnote{In biology a gene-pool is all the genes in a population \cite{lawrence1989hsd}, but here it is the set of genes $\{g_{i}, ..., g_{j}\}$ available at a habitat $h$.}, $g_{p}$, of the user's habitat $h$, and from there the migration of the gene $g$ occurs, which involves migrating (copying) the gene $g$ probabilistically ($p(h)>0$) to all the connected habitats $h$. The gene $g$ is copied rather than moved, because the gene $g$ may also be of use to the providing user $u$. The copying of a gene $g$ to a connected habitat $h$ depends on the associated migration probability. If the probability were one ($p(h)=1$), then it would definitely be sent. When migration occurs, depending on the probabilities associated with the habitat $h$ connections $c$, an exact copy of the gene $g$ is made at a connected habitat $h$. The successful use of the migrated gene $g$, in response to user requests for applications $r$, will lead to further migration (distribution) and therefore availability of the gene $g$ to other users $u$.

So, the connections $c$ joining the habitats $h$ are reinforced by successful gene $g$ migration. The success of the migration, the \emph{migration feedback}, leads to the reinforcing and creation of migration links between the habitats $c$, just as the failure of migration leads to the weakening and negating of migration links between the habitats $h$. The success of migration is determined by the usage of the genes $g$ at the habitats $h$ to which they migrate. When a gene-set $\{g_{i}, ..., g_{j}\}$ is found and used in responding to a user request, then the appropriate connection probabilities can be updated. If the gene-set $\{g\}$ was fully or partly evolved elsewhere, then where the set or subsets were created needs to be passed on to the connection probabilities, because the value in a gene-set $\{g\}$ is the unique combination it provides of the individual genes $g$ contained within. So, it is necessary to manage the feedback to the connection probabilities for migrating gene-sets $\{g_{i}, ..., g_{j}\}$, and not just the individual genes $g$ contained within the sequence, including the partial use of a gene-set, $\{g\}$, in a newly evolved one, $\{g_{i}, ..., g_{j}\}$. Specifically, the mechanism for \emph{migration feedback} needs to know the habitats $h$ where migrating gene-sets $\{g\}$ were created, to create new connections or reinforce existing connections to these habitats $h$. The global effect of the gene $g$ migration and \emph{migration feedback} on the habitat network $H$ is the clustering of habitats $h$ around the communities present within the user base $U$.

\tfigure{width=0.9\columnwidth}{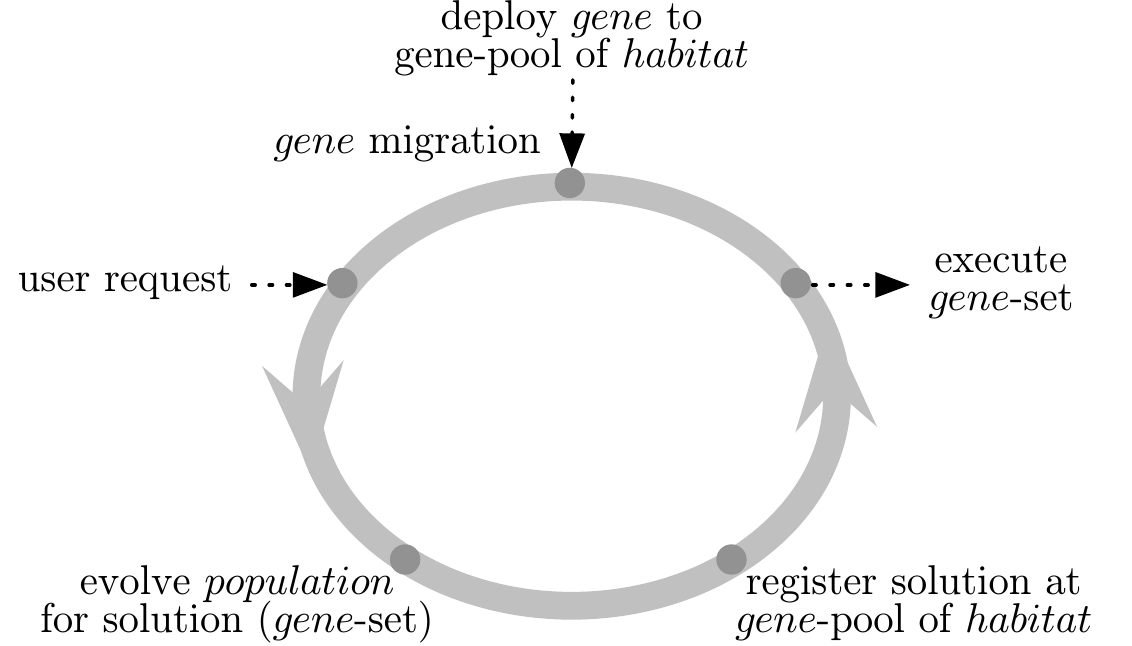}{pdf}{Gene Life-Cycle}{Begins with deployment to its users's habitat $h$ for distribution within the habitat network $H$. It can then be used in evolving the optimal gene-sets(s) in response to a user request $r$. The optimal gene-set $\{g_{i}, ..., g_{j}\}$ is then registered at the habitat $h$. If a gene-set solution $\{g\}$ is then executed, an attempt is made to migrate (copy) it to every other connected habitat $h$, success depending on the probability associated with the connection.}{-1mm}{!b}{-2mm}{}

So, a gene's life-cycle begins with deployment to its users's habitat $h$ for distribution within the habitat network $H$. The gene $g$ is then migrated to any habitats connected to their user's habitat ($p(h_{i})>0$), to make it available in other habitats where it could potentially be useful. The gene $g$ is then available to the local evolutionary optimisation, to be used in evolving the optimal gene-sets(s) in response to a user request $r$. The optimal gene-set $\{g_{i}, ..., g_{j}\}$ is then registered at the habitat $h$, being stored in the habitat's gene-pool $g_{p}$. If a gene-set solution $\{g\}$ is then executed, an attempt is made to migrate (copy) it to every other connected habitat $h$, success depending on the probability associated with the connection. A gene $g$ can also be deleted if after several successive user requests at a habitat $h$ it remains unused; it will have a small number of \emph{escape migrations}, in which it is not copied, but is randomly moved to another connected habitat $h$. If the gene $g$ fails to find a \emph{niche} before running out of \emph{escape migrations}, then it will be deleted.

The \emph{escape range} is the number of escape migrations available to a gene $g$ upon the risk of death (deletion). If a gene $g$ migrates to a habitat $h$ and is not used after several user requests $r$, then it will have the opportunity to migrate (move not copy) randomly to another connected habitat $h$. After this happens several times the gene $g$ will be deleted (die). The \emph{escape range} will be dynamically responsive to the size of the habitat $h$ cluster that the gene $g$ exists within. This creates a dynamic \emph{time-to-live} \cite{comer1988iti} for the genes $g$, such that those that are used more will live longer and distribute further than those that are used less.

\subsection{Population}

A population in a natural ecosystem is all the members of a species that occupy a particular area at a given time \cite{lawrence1989hsd}. Our population, $P$, is also \emph{all the members of a species that occupy a particular area at a given time}, like an island from the island-models of \acl{DEC} \cite{lin1994cgp}. So, the population $P$ represents evolving populations in our ecosystem-oriented distributed evolutionary computing. The population $P$ is instantiated to evolve an optimal algorithm (set of genes $\{g_{i}, ..., g_{j}\}$) to an individual user request $r$, from the set of genes (gene-pool, $g_{p}=\{g_{1}, ..., g_{j}\}$) available at the habitat $h$ where it is instantiated. The solution is found by evolving the population of solutions $[\{g_{i}, ..., g_{j}\}, ..., \{g_{k}, ..., g_{l}\}]$, with the evolutionary process of a population $P$ providing a \emph{combinatorial optimisation} \cite{papadimitriou1998coa} of the genes $g$ available, when responding to a user request $r$. We shall assume that the genes are not mutated as they are atomic by definition. So, as in genetic algorithms \cite{goldberg}, \emph{mutations} will occur by switching genes $g$ in and out of the gene-sets $\{g\}$ that define individuals. \emph{Recombination} (crossover) will occur by performing a crossing of two individuals (gene-sets). Therefore, an evolving population $P$ consists of the selection pressure $s$ defined by the user $u$ request $r$, initialised with the gene-pool $\{g\}$ of the habitat $h$ where it is instantiated. Then, the evolving population $P$ becomes a collection of gene-sets $\{g\}$ evolving to the selection pressure $s$ created from the user request $r$,
\begin{equation}
population = P(r,g_{p},g_{s})=[\{g_{i}, ..., g_{j}\}, ..., \{g_{k}, ..., g_{l}\}].
\end{equation}

The evolutionary optimisation is accelerated in the following three ways: first, the habitat $h$ provides an optimal subset of the genes $g$ available globally $G$ in the ecosystem $E$, which is localised to the specific user $u$ it represents; second, making use of gene-sets, $\{g_{i}, ..., g_{j}\}$, previously evolved in response to the user's earlier requests $r_{e}$; and third, taking advantage of relevant gene-sets evolved elsewhere in response to similar requests $r_{s}$ by other users. The population $P$ then proceeds to evolve the optimal gene-set(s) $\{g\}$ that fulfils the user request $r$, and as the genes $g$ are the base unit for evolution, it searches the available gene-set combination space $\mathcal{P}(g_{p})$. For an evolved gene-set $\{g\}$ that is executed (instantiated) by the user $u$, it then migrates to other peers (habitats $h$) becoming hosted where it is useful, to combine with other genes $g$ in other populations $P$ to assist in responding to other user requests for algorithms $r$.

There is a huge body of work and continuing research regarding theoretical approaches to evolutionary computing \cite{eiben2003iec}, including the extensive use of genetic algorithms for practical real-world problem solving \cite{ducheyne2003fiu}. Full use should make of the current state-of-the-art, and future developments in the area of evolutionary computing \cite{jin2005csf} in determining the optimal evolutionary computing approaches to have in the populations $P$ of our ecosystem-oriented distributed evolutionary computing $E$.

\subsection{Habitat}

The habitats $h$ are functionally analogous to the habitats of a biological ecosystem \cite{lawrence1989hsd}, providing a distributed environment for gene migration and evolution to occur, extending the \emph{island-model} of \acl{DEC} \cite{lin1994cgp} for the connectivity between habitats $h$. There will be a habitat $h$ for each user $u$, through which they will submit their requests $r$ for algorithms. So, the habitat $h$ represents a location at which a gene $g$ can move to, and so the collection of genes $g$ at the habitat $h$ defines the gene-pool $g_{p}$ with which to seed populations $P$ to solve user requests $r$ for algorithms. Supporting this functionality, habitats $h$ have the following core functions:
\begin{itemize}
\item Provide a subset of the genes $g$ available globally $G$, relevant to the user $u$ that it represents, the gene-pool $g_{p}$.
\item Accelerate, via its gene-pool $g_{p}$, the populations $P$ instantiated to evolve optimal gene-sets $\{g_{i}, ..., g_{j}\}$ in response to user requests for algorithms $r$.
\item Manage the inter-habitat connections for gene migration, $c=\{p(h_{1}), ..., p(h_{i})\}$.
\end{itemize}

The collection of genes $g_{p}$ at each habitat $h$ (peer) will change over time, as the more successful genes $g$ spread through the habitat network, $H=\{h_{1}, ..., h_{i}\}$, of the ecosystem $E$, and as the less successful genes $g$ are deleted. Successive user requests $r$ over time $t$ to their dedicated habitats $h$ makes this process possible, because the continuous and varying user requests for applications, $R=\{r_{i}, ..., r_{j}\}$, provide a dynamic evolutionary pressure, $S=\{s_{i}, ..., s_{j}\}$, on the genes globally $G=\{g_{i}, ..., g_{j}\}$, which have to evolve to better satisfy those requests. So, the genes $g$ will recombine and evolve over time, constantly seeking to increase their effectiveness for the user base $U=\{u_{1}, ..., u_{i}\}$. Additionally, it also stores gene-sets, $g_{s}$ evolved from the habitat's populations $P$, and gene-sets that migrate to the habitat $h$ from other users' habitats, because they can potentially accelerate future populations $P$ instantiated to respond to user requests $r$. So, a habitat consists of the connections to other habitats $c=\{p(h_{i}), ..., p(h_{j})\}$, the genes available (gene-pool $g_{p}$), previously evolved gene-sets $g_{s}$ available, and current populations $P$ being evolved,
\begin{equation}
habitat = h = (c, g_{p}, g_{s}, \{P_{i}, ..., P_{j}\}).
\end{equation}

The landscape, in energy-centric biological ecosystems, defines the connectivity between habitats \cite{begon96}. Connectivity of nodes in the digital world is generally not defined by geography or spatial proximity, but by information or semantic proximity. For example, connectivity in a peer-to-peer network is based primarily on bandwidth and information content, and not geography. The island-models of \acl{DEC} use an information-centric model for the connectivity of nodes (\emph{islands}) \cite{lin1994cgp}. However, because it is generally defined for one-time use (to evolve a solution to one problem and then stop) it usually has a fixed connectivity between the nodes, and therefore a fixed topology \cite{cantupaz1998spg}. So, supporting ecosystem-oriented distributed evolutionary computing, with dynamic multi-objective \emph{selection pressures} (fitness landscapes \cite{wright1932} with many peaks), requires a re-configurable network topology, such that habitat connectivity $c$ can be dynamically adapted based on the observed migration paths of the genes $g$ between the users $u$ within the habitat network $H$. So, based on the island-models of \acl{DEC} \cite{lin1994cgp}, each connection between the habitats $h$ is bi-directional and there is a probability associated with moving in either direction across the connection, with the connection probabilities $c$ affecting the rate of migration of the genes. However, additionally, the connection probabilities will be updated by the success or failure of gene migration using the concept of Hebbian learning \cite{hebb}: the habitats $h$ which do not successfully exchange genes $g$ will become less strongly connected, and the habitats $h$ which do successfully exchange genes $g$ will achieve stronger connections. This leads to a topology that adapts over time, resulting in a network that supports and resembles the connectivity of the user base $U$. We will further discuss a resulting topology in the next section.

When a new user $u$ joins, a habitat $h$ needs to be created for them, and most importantly connected to the correct cluster(s) within the habitat network $H=\{h_{1}, ..., h_{i}\}$ of the ecosystem $E$. A new user's habitat $h_{u}$ can be connected randomly to the habitat network $H$, as it will dynamically reconnect based upon the user's behaviour. User profiling can also be used to help connect a new user's habitat $h_{u}$ to the optimal part of the network $H$, by finding a similar user $u_{s}$ or asking the user to identify a similar user $u_{s}$, and then cloning their habitat's $h_{u_{s}}$ connections $c$ (the set of probabilities for connecting to other habitats). Also, when a new habitat $h$ is created, its gene-pool $g_{p}$ should be created by merging the gene-pools of the habitats to which it is initially connected.

\subsection{Ecosystem}
Our ecosystem-oriented distributed evolutionary computing supports the automatic re-combining of numerous genes, $g$, simple atomic algorithms, by their interaction in evolving populations, $P$, to meet user, $u$, requests, $r$, for algorithms, in a scalable architecture of distributed interconnected habitats, $h$. The sharing of genes $g$ between habitats $h$ ensures the system is scalable, while maintaining a high evolutionary specialisation for each user, $u$. The network of interconnected habitats, $H$, is equivalent to the \emph{abiotic} environment of biological ecosystems \cite{begon96}; combined with the genes $g$ within the populations $P$ and the gene migration for \acl{DEC}, with the environmental selection pressures $S$ provided by a user base $\{u\}$, then the union of the habitats $h$ creates the ecosystem $E$. The continuous and varying user requests for algorithms provide a dynamic evolutionary pressure on the populations of genes, which have to evolve to better fulfil those user requests. So, local and global optimisations concurrently operate to determine solutions to satisfy different optimisation problems. Therefore, the local search is improved through this twofold process to yield better local optima faster, as the distributed optimisation provides prior sampling of the search space through computations already performed in other peers with similar constraints. So, an ecosystem can be defined as the habit network $H=\{h_{1}, ..., h_{i}\}$, the user base $U=\{u_{1}, ..., u_{i}\}$, the requests they submit $R=\{r_{i}, ..., r_{j}\}$ which provides a dynamic evolutionary pressure $S=\{s_{i}, ..., s_{j}\}$,
\begin{equation}
ecosystem = E = (H, U, R, S).
\end{equation}

\section{Topology}
\label{topology} 

Our ecosystem-oriented form of \acl{DEC} allows for the connectivity of the habitats $h$ to adapt to the connectivity within the user base $U$, with a cluster of habitats representing a community within the user base. If a user $u$ is a member of more than one community, the user's habitat $h_{u}$ will be in more than one cluster. This leads to a network topology that will be discovered with time, and which reflects the connectivity within the user base $U$. Similarities in requests by different users will reinforce behavioural patterns, and lead to clustering of the habitats within the ecosystem, which can occur over geography, language, etc. This will form communities for more effective information sharing, the creation of niches, and will improve the responsiveness of the system. 

The connections between the habitats are self-managed, through the migration between the habitats. Essentially, successful migration will reinforce habitat connections, thereby increasing the probability of future migration along these connections. If a successful multi-hop migration occurs, then a new link between the start and end habitats can be formed. Unsuccessful migrations will lead to connections (migration probabilities) decreasing, until finally the connection is closed. So, connections between habitats $h$ are reconfigured depending on the connectivity of the user base $U$, the habitat clustering will therefore be parallel to the user communities, as shown in Figure \ref{DBE}. The communities will cluster over defining properties of the user base, such as language, nationality, geography, etc. So, the system will take on a topology similar to that of the user base.

\tfigure{width=0.9\columnwidth}{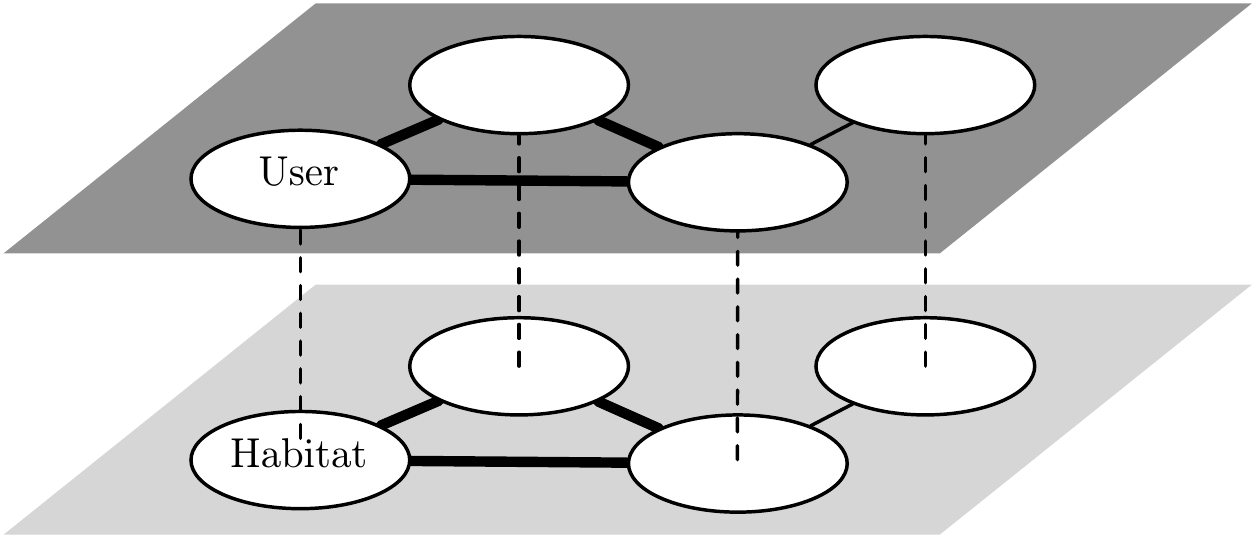}{graffle}{Ecosystem-Oriented Distributed Evolutionary Computing - User Connectivity}{As the connections between habitats are reconfigured depending on the connectivity of the user base, the habitat clustering will therefore be parallel to the user communities.}{-2mm}{}{}{}

Landscape connectivity plays an important part in ecosystems. When the density of habitats within an environment falls below a critical threshold, widespread species may fragment into isolated populations. Fragmentation can have several consequences. Within populations, these effects include loss of genetic diversity and detrimental inbreeding \cite{GreenKirley}. At a broader scale, isolated populations may diverge genetically, leading to speciation. Fragmentation of the habitat network $H$ could occur, but only if dictated by the structure of the user base $U$. The issue of greater concern is when individual habitats $h$ become totally disconnected, which could only occur under certain conditions. One condition is that the evolved solutions (algorithms) consistently fail to satisfy user requests $r$. Another condition is when the solutions they share are undesirable to the users that are within the migration range of these solutions. These scenarios could arise because the habitat $h$ is located within the wrong cluster, in which case the user can be asked to join another cluster within the habitat network, assuming the user base $U$ is of sufficient size to provide a viable alternative.

\section{Discussion}
\label{discussion}
Ecosystem-oriented \acl{DEC} proved significantly different to natural ecosystems with respect to the information-centric dynamically re-configurable network topology. A re-configurable network topology is required to support the constantly changing multi-objective information-centric \emph{selection pressures} $S$ of the user base $U$. Hence, using the concept of Hebbian learning \cite{hebb}, habitat connectivity is dynamically adapted based on the observed migration paths of the genes within the habitat network. We would argue that this difference is not compromises, but a unique feature that arrises from mimicking natural ecosystems. Biomimicry is best when not slavish imitation; it is inspiration using the principles which nature has demonstrated to be successful design strategies \cite{biomimicry}. So while the topology of natural ecosystems are energy-centric, the topology of ecosystem-oriented \acl{DEC} is information-centric.

Our ecosystem-oriented \acl{DEC} proposes a fundamental paradigm shift, from a \emph{pull}-oriented approach to a \emph{push}-oriented approach. So, instead of the \emph{pull}-oriented approach of generating algorithms only upon request \cite{singh2005soc}, a \emph{push}-oriented approach is followed of distributing and composing algorithms pre-emptively, as well as upon request. Given the increasing need to tailor software to specific uses and users, and the need to deal with increasing software complexity, we consider this shift from pull to push will become paramount.

We have previously considered a case study, Digital Business Ecosystems \cite{abcdbe, thesis}, where the genes $g$ where mapped to a hybrid construct of agents and services to create a Digital Ecosystem \cite{bionetics, thesis} capable of supporting a business ecosystem (a networks of \aclp{SME}). This included experimental simulation results, comparing to service-oriented computing \cite{bionetics} and natural ecosystems \cite{de07oz, thesis}. We also considered an abstraction of this Digital Ecosystem, based within \aclp{CAS}, for applicability to other classes of systems that could be considered as, or augmented to, 'eco'-systems \cite{caes}.

\section{Conclusion}
\label{conclusion}

We have defined the fundamentals for a new class of system, ecosystem-oriented distributed evolutionary computing, created through combining understanding from theoretical ecology, evolutionary theory, and \acl{DEC}. By comparing and contrasting the relevant theoretical ecology we have examined how some ecological features may emerge in distributed evolutionary computing. Specifically, we suggested that ecosystem-orientated distributed evolutionary computing, like natural ecosystems, will usually consist of self-replicating individuals that interact both with one another and with an external environment \cite{begon96}. Population dynamics and evolution, and spatial and network interactions, will all influence the behaviour of these systems. Many of these properties can be understood via well-known ecological models \cite{MacArthur,Hubbell}. These models provide a theoretical basis for the occurrence of self-organisation, in artificial and biological ecosystems, resulting from the interactions among the individuals and their environment, leading to complex non-linear behaviour \cite{MacArthur,Hubbell,Levin}; and it is this property that provides the underlying potential for scalable problem-solving in artificial environments.

We have strived such that the word \emph{ecosystem} is more than just a metaphor, instead creating an artificial counterpart of natural ecosystems, and therefore potentially having their desirable properties, such as scalability and self-organisation. The stability, convergence and equilibrium of the evolving populations of our ecosystem-orientated distributed evolutionary computing have been analysed here \cite{stabEvo, agentStability}, while their complexity and self-organisation have been considered here \cite{phyComSym,phycom,acm}, and the overall diversity of the ecosystem has been considered here \cite{eoa}. This, and other work, will help to provide theoretical bounds on the control parameters. There is also an obviously desirability to accelerate and optimise our ecosystem-oriented \acl{DEC}, for which we have explored the application of augmentations that interact with the ecosystem dynamics \cite{de08th}.

The ever-increasing challenge of software complexity in creating progressively more sophisticated and distributed applications, makes the design and maintenance of efficient and flexible systems a growing challenge \cite{newsArticle3}, for which current software development techniques have hit a \emph{complexity wall} \cite{lyytinen2001nwn}. In response we have created ecosystem-oriented \acl{DEC}, possessing properties of self-organisation, scalability and sustainability from ecosystems \cite{Levin}. So, overcoming the challenge by allowing the automated search for new algorithms in a scalable architecture, through the evolution of software in a distributed network.

\bibliographystyle{IEEEtran}
\bibliography{references,myRefs}

\begin{thebibliography}{10}
\providecommand{\url}[1]{#1}
\csname url@samestyle\endcsname
\providecommand{\newblock}{\relax}
\providecommand{\bibinfo}[2]{#2}
\providecommand{\BIBentrySTDinterwordspacing}{\spaceskip=0pt\relax}
\providecommand{\BIBentryALTinterwordstretchfactor}{4}
\providecommand{\BIBentryALTinterwordspacing}{\spaceskip=\fontdimen2\font plus
\BIBentryALTinterwordstretchfactor\fontdimen3\font minus
  \fontdimen4\font\relax}
\providecommand{\BIBforeignlanguage}[2]{{%
\expandafter\ifx\csname l@#1\endcsname\relax
\typeout{** WARNING: IEEEtran.bst: No hyphenation pattern has been}%
\typeout{** loaded for the language `#1'. Using the pattern for}%
\typeout{** the default language instead.}%
\else
\language=\csname l@#1\endcsname
\fi
#2}}
\providecommand{\BIBdecl}{\relax}
\BIBdecl

\bibitem{ecpaper}
P.~Marrow, ``Nature-inspired computing technology and applications,'' \emph{BT
  Technology Journal}, vol.~18, pp. 13--23, 2000.

\bibitem{ec15}
C.~Darwin, \emph{On the Origin of Species by Means of Natural Selection}.\hskip
  1em plus 0.5em minus 0.4em\relax John Murray, 1859.

\bibitem{ec16}
D.~Futuyma, \emph{Evolutionary Biology}.\hskip 1em plus 0.5em minus 0.4em\relax
  Sinauer Associates, 1998.

\bibitem{ec17}
T.~Baeck, D.~Fogel, and Z.~Michalewicz, Eds., \emph{Handbook of Evolutionary
  Computation}.\hskip 1em plus 0.5em minus 0.4em\relax CRC Press, 1997.

\bibitem{eiben2003iec}
A.~Eiben and J.~Smith, \emph{Introduction to Evolutionary Computing}.\hskip 1em
  plus 0.5em minus 0.4em\relax Springer, 2003.

\bibitem{Levin}
S.~Levin, ``Ecosystems and the biosphere as complex adaptive systems,''
  \emph{Ecosystems}, vol.~1, pp. 431--436, 1998.

\bibitem{cantupaz1998spg}
E.~Cantu-Paz, ``A survey of parallel genetic algorithms,'' \emph{R{\'e}seaux et
  syst{\`e}mes r{\'e}partis, Calculateurs Parall{\`e}les}, vol.~10, pp.
  141--171, 1998.

\bibitem{stender1993pga}
J.~Stender, \emph{Parallel Genetic Algorithms: Theory and Applications}.\hskip
  1em plus 0.5em minus 0.4em\relax IOS Press, 1993.

\bibitem{devisser2007ese}
J.~de~Visser and S.~Elena, ``The evolution of sex: empirical insights into the
  roles of epistasis and drift,'' \emph{Nature Reviews Genetics}, vol.~8, pp.
  139--49, 2007.

\bibitem{lin1994cgp}
S.~Lin, W.~Punch~III, and E.~Goodman, ``Coarse-grain parallel genetic
  algorithms: categorization and new approach,'' in \emph{Symposium on Parallel
  and Distributed Processing}.\hskip 1em plus 0.5em minus 0.4em\relax IEEE
  Press, 1994, pp. 28--37.

\bibitem{levins1969sda}
R.~Levins, ``Some demographic and genetic consequences of environmental
  heterogeneity for biological control,'' \emph{Bulletin of the Entomological
  Society of America}, vol.~15, pp. 237--240, 1969.

\bibitem{begon96}
M.~Begon, J.~Harper, and C.~Townsend, \emph{Ecology: Individuals, Populations
  and Communities}.\hskip 1em plus 0.5em minus 0.4em\relax Blackwell
  Publishing, 1996.

\bibitem{yuchi2008}
M.~Yuchi and J.~Kim, ``{Ecology-inspired evolutionary algorithm using
  feasibility-based grouping for constrained optimization},'' in \emph{Congress
  on Evolutionary Computation}, vol.~2, 2008, pp. 1455--1461.

\bibitem{hoile2002core}
C.~Hoile, F.~Wang, E.~Bonsma, and P.~Marrow, ``{Core specification and
  experiments in DIET: a decentralised ecosystem-inspired mobile agent
  system},'' in \emph{Proceedings of the first international joint conference
  on Autonomous agents and multiagent systems: part 2}.\hskip 1em plus 0.5em
  minus 0.4em\relax ACM, 2002, pp. 623--630.

\bibitem{epi}
G.~Briscoe and P.~{De Wilde}, ``The computing of digital ecosystems,''
  \emph{International Journal of Organizational and Collective Intelligence},
  vol.~1, no.~4, pp. 1--17, 2010.

\bibitem{acmMedes}
------, ``Computing of applied digital ecosystems,'' in \emph{ACM Management of
  Emergent Digital Ecosystems Conference}, 2009, pp. 28--35.

\bibitem{thesis}
G.~Briscoe, ``Digital ecosystems,'' Ph.D. dissertation, Imperial College
  London, 2009.

\bibitem{dbebkpub}
G.~Briscoe and S.~Sadedin, ``Natural science paradigms,'' in \emph{Digital
  {B}usiness {E}cosystems}.\hskip 1em plus 0.5em minus 0.4em\relax European
  {C}ommission, 2007, pp. 48--55.

\bibitem{de07oz}
G.~Briscoe, S.~Sadedin, and G.~Paperin, ``Biology of applied digital
  ecosystems,'' in \emph{IEEE Digital Ecosystems and Technologies Conference},
  2007, pp. 458--463.

\bibitem{bionetics}
G.~Briscoe and P.~{De Wilde}, ``Digital {E}cosystems: Evolving service-oriented
  architectures,'' in \emph{IEEE Bio Inspired Models of Network, Information
  and Computing Systems Conference}, 2006, pp. 1--6, (Most Cited Paper in The
  Field).

\bibitem{longman}
A.~Redmore and M.~Griffin, \emph{Advanced Level and Advanced Special Level
  Biology}.\hskip 1em plus 0.5em minus 0.4em\relax Longman Group Limited, 1994.

\bibitem{lawrence1989hsd}
E.~Lawrence, \emph{Henderson's dictionary of biological terms}.\hskip 1em plus
  0.5em minus 0.4em\relax Pearson Education, 2005.

\bibitem{hanski1999me}
I.~Hanski, \emph{Metapopulation Ecology}.\hskip 1em plus 0.5em minus
  0.4em\relax Oxford University Press, 1999.

\bibitem{hanski2003mtf}
I.~Hanski and O.~Ovaskainen, ``Metapopulation theory for fragmented
  landscapes,'' \emph{Theoretical Population Biology}, vol.~64, pp. 119--127,
  2003.

\bibitem{Greenetal2006}
D.~Green, N.~Klomp, G.~Rimmington, and S.~Sadedin, \emph{Complexity in
  Landscape Ecology}.\hskip 1em plus 0.5em minus 0.4em\relax Springer, 2006.

\bibitem{suzie}
D.~Green and S.~Sadedin, ``Interactions matter- complexity in landscapes and
  ecosystems,'' \emph{Ecological Complexity}, vol.~2, pp. 117--130, 2005.

\bibitem{Bell}
G.~Bell, ``The distribution of abundance in neutral communities,''
  \emph{American Naturalist}, vol. 396, pp. 606--617, 2000.

\bibitem{sizling2004pls}
A.~Sizling and D.~Storch, ``Power-law species-area relationships and
  self-similar species distributions within finite areas,'' \emph{Ecology
  Letters}, vol.~7, pp. 60--68, 2004.

\bibitem{Hubbell}
S.~Hubbell, \emph{The Unified Neutral Theory of Biodiversity and
  Biogeography}.\hskip 1em plus 0.5em minus 0.4em\relax Princeton University
  Press, 2001.

\bibitem{Sole}
R.~Sol{\'e}, D.~Alonso, and A.~McKane, ``Self-organized instability in complex
  ecosystems,'' \emph{Philosophical Transactions: Biological Sciences}, vol.
  357, pp. 667--681, 2002.

\bibitem{Greenetal2000}
D.~Green, D.~Newth, and M.~Kirley, ``Connectivity and catastrophe - towards a
  general theory of evolution,'' in \emph{International Conference on
  Artificial Life}, M.~Bedau, Ed.\hskip 1em plus 0.5em minus 0.4em\relax MIT
  Press, 2000, pp. 153--161.

\bibitem{Greenetalinpress}
D.~Green, T.~Leishman, and S.~Sadedin, ``Dual phase evolution: a mechanism for
  self-organization in complex systems,'' in \emph{International Conference on
  Complex Systems}, A.~Minai, D.~Braha, and Y.~Bar-Yam, Eds.\hskip 1em plus
  0.5em minus 0.4em\relax Springer, 2006.

\bibitem{detoro2002ppg}
F.~de~Toro, J.~Ortega, J.~Fernandez, and A.~Diaz, ``{PSFGA}: a parallel genetic
  algorithm for multiobjective optimization,'' in \emph{Euromicro Workshop on
  Parallel, Distributed and Network-based Processing}, F.~Vajda and
  N.~Podhorszki, Eds.\hskip 1em plus 0.5em minus 0.4em\relax IEEE Press, 2002,
  pp. 384--391.

\bibitem{back1996eat}
T.~B{\"a}ck, \emph{Evolutionary Algorithms in Theory and Practice: Evolution
  Strategies, Evolutionary Programming, Genetic Algorithms}.\hskip 1em plus
  0.5em minus 0.4em\relax Oxford University Press, 1996.

\bibitem{papadimitriou1998coa}
C.~Papadimitriou and K.~Steiglitz, \emph{Combinatorial Optimization: Algorithms
  and Complexity}.\hskip 1em plus 0.5em minus 0.4em\relax Dover Publications,
  1998.

\bibitem{comer1988iti}
D.~Comer, \emph{Internetworking with TCP/IP: principles, protocols, and
  architecture}.\hskip 1em plus 0.5em minus 0.4em\relax Prentice Hall, 2005.

\bibitem{goldberg}
D.~Goldberg, \emph{Genetic algorithms in search, optimization, and machine
  learning}.\hskip 1em plus 0.5em minus 0.4em\relax Addison-Wesley, 1989.

\bibitem{ducheyne2003fiu}
E.~Ducheyne, B.~De~Baets, and R.~De~Wulf, ``Is fitness inheritance useful for
  real-world applications,'' in \emph{Evolutionary Multi-criterion
  Optimization}, C.~Fonseca, Ed.\hskip 1em plus 0.5em minus 0.4em\relax
  Springer, 2003, pp. 31--42.

\bibitem{jin2005csf}
Y.~Jin, ``A comprehensive survey of fitness approximation in evolutionary
  computation,'' \emph{Soft Computing - A Fusion of Foundations, Methodologies
  and Applications}, vol.~9, pp. 3--12, 2005.

\bibitem{wright1932}
S.~Wright, ``The roles of mutation, inbreeding, crossbreeding and selection in
  evolution,'' in \emph{International Congress on Genetics}, D.~Jones,
  Ed.\hskip 1em plus 0.5em minus 0.4em\relax Brooklyn botanic garden, 1932, pp.
  356--366.

\bibitem{hebb}
D.~Hebb, \emph{The Organization of Behavior}.\hskip 1em plus 0.5em minus
  0.4em\relax Wiley, 1949.

\bibitem{GreenKirley}
D.~Green and M.~Kirley, ``Adaptation, diversity and spatial patterns,''
  \emph{International Journal of Knowledge-Based Intelligent Engineering
  Systems}, vol.~4, pp. 184--190, 2000.

\bibitem{biomimicry}
J.~Benyus, \emph{Biomimicry, Innovation Inspired by Nature}.\hskip 1em plus
  0.5em minus 0.4em\relax Harper Collins Publishers, 2002.

\bibitem{abcdbe}
J.~Stanley and G.~Briscoe, ``The abc of digital business ecosystems,''
  \emph{Communications Law - Journal of Computer, Media and Telecommunications
  Law}, vol.~15, no.~1, 2010.

\bibitem{caes}
G.~Briscoe, ``Complex adaptive digital ecosystems,'' in \emph{ACM Management of
  Emergent Digital Ecosystems Conference}, 2010, pp. 39--46.

\bibitem{singh2005soc}
M.~Singh and M.~Huhns, \emph{Service-Oriented Computing: Semantics, Processes,
  Agents}.\hskip 1em plus 0.5em minus 0.4em\relax Wiley, 2005.

\bibitem{MacArthur}
R.~MacArthur and E.~Wilson, \emph{The Theory of Island Biogeography}.\hskip 1em
  plus 0.5em minus 0.4em\relax Princeton University Press, 1967.

\bibitem{stabEvo}
P.~{De Wilde} and G.~Briscoe, ``Stability of evolving multiagent systems,''
  \emph{Systems, Man and Cybernetics - Part B}, vol.~41, no.~4, pp. 1149--1157,
  2011.

\bibitem{agentStability}
G.~Briscoe and P.~{De Wilde}, ``Digital {E}cosystems: Stability of evolving
  agent populations,'' in \emph{ACM Management of Emergent Digital Ecosystems
  Conference}, 2009, pp. 36--43.

\bibitem{phyComSym}
------, ``Physical complexity of variable length symbolic sequences,''
  \emph{Physica A}, vol. 390, pp. 3732--3741, 2011.

\bibitem{phycom}
------, ``Digital {E}cosystems: {S}elf-organisation of evolving agent
  populations,'' in \emph{ACM Management of Emergent Digital Ecosystems
  Conference}, 2009, pp. 44--48.

\bibitem{acm}
------, ``Self-organisation of evolving agent populations in digital
  ecosystems,'' \emph{International Journal of Internet and Enterprise
  Management}, vol.~7, no.~3, pp. 239--286, 2011.

\bibitem{eoa}
G.~Briscoe, S.~Sadedin, and P.~{De Wilde}, ``Digital ecosystems:
  Ecosystem-oriented architectures,'' \emph{Natural Computing}, vol.~10, no.~3,
  pp. 1143--1194, 2011.

\bibitem{de08th}
G.~Briscoe and P.~{De Wilde}, ``Digital {E}cosystems: {O}ptimisation by a
  distributed intelligence,'' in \emph{IEEE Digital Ecosystems and Technologies
  Conference}, 2008, pp. 192--197, (Awarded Best Track Paper).

\bibitem{newsArticle3}
\BIBentryALTinterwordspacing
J.~Markoff, ``Faster chips are leaving programmers in their dust,'' New York
  Times, Tech. Rep., 2007. [Online]. Available:
  \url{http://www.nytimes.com/2007/12/17/technology/17chip.html}
\BIBentrySTDinterwordspacing

\bibitem{lyytinen2001nwn}
K.~Lyytinen and Y.~Yoo, ``The next wave of nomadic computing: A research agenda
  for information systems research,'' \emph{Sprouts: Working Papers on
  Information Systems}, vol.~1, pp. 1--20, 2001.

\end{thebibliography}

\end{document}